\RequirePackage{amsmath}
\documentclass{llncs}
\usepackage{graphicx}
\usepackage{rotating}
\usepackage[T1, T2A]{fontenc}
\usepackage{hyperref}
\graphicspath{ {figs/} }

\DeclareMathOperator*{\argmin}{arg\,min}
\usepackage{filecontents}

\begin{document}
\title{Exploring Uncertainty Measures in Deep Networks for Multiple Sclerosis Lesion Detection and Segmentation}
\titlerunning{Exploring Uncertainty} %

\author{Tanya Nair\inst{1}, Doina Precup\inst{2}, Douglas L. Arnold\inst{3,4}, Tal Arbel\inst{1}}
\authorrunning{Nair, T. et al.} 
%
\institute{Centre for Intelligent Machines, McGill University, Montr\'eal, Canada,\\
\email{tnair@cim.mcgill.ca},\\
\and
School of Computer Science, McGill University, Montr\'eal, Canada,
\and
Montreal Neurological Institute, McGill University,  Montr\'eal, Canada,
\and
NeuroRx Research, Montr\'eal, Canada}

\maketitle              

\begin{abstract}
Deep learning (DL) networks have recently been shown to outperform other segmentation methods on various public, medical-image challenge datasets \cite{carass2017longitudinal,menze2015multimodal,styner20083d}, especially for large pathologies. However, in the context of diseases such as Multiple Sclerosis (MS), monitoring all the focal lesions visible on MRI sequences, even very small ones, is essential for disease staging, prognosis, and evaluating treatment efficacy. Moreover, producing deterministic outputs hinders DL adoption into clinical routines. Uncertainty estimates for the predictions would permit subsequent revision by clinicians. 
We present the first exploration of multiple uncertainty estimates based on Monte Carlo (MC) dropout~\cite{gal2016dropout} in the context of deep networks for lesion detection and segmentation in medical images. Specifically, we develop a 3D MS lesion segmentation CNN, augmented to provide four different voxel-based uncertainty measures based on MC dropout.
We train the network on a proprietary, large-scale, multi-site, multi-scanner, clinical MS dataset, and compute lesion-wise uncertainties by accumulating evidence from voxel-wise uncertainties within detected lesions. We analyze the performance of voxel-based segmentation and lesion-level detection by choosing operating points based on the uncertainty. Empirical evidence suggests that uncertainty measures consistently allow us to choose superior operating points compared only using the network's sigmoid output as a probability.

\end{abstract}

\section{Introduction}
Deep learning (DL) has become ubiquitous in computer vision and other applications~\protect{\cite{hernandez2016team,russakovsky2015imagenet}}, yet its adoption in medical imaging has been comparatively slow, due, in part, to the shortage of large-scale annotated datasets. Recently, DL frameworks have been shown to outperform other segmentation methods on a variety of public challenge datasets~\cite{carass2017longitudinal,menze2015multimodal,styner20083d}, particularly on metrics focused on large pathologies. For neurological diseases such as Multiple Sclerosis (MS), lesions can be very small (eg. 3-5 voxels) and the detection and segmentation of lesions of {\it all sizes} on MRI sequences is a key component for clinical assessment of disease stage and prognosis, as well as for evaluating the efficacy of treatments during clinical trials. Early DL approaches have shown success in the segmentation of large lesions~\cite{brosch2016deep}, and recent work has shown how using a tissue-prior or lesion-prior can improve detection for medium and small lesions on small, private datasets~\cite{ghafoorian2016non}. However, current DL methods have not yet been shown to outperform other machine learning methods in the detection of small lesions, leading to potential errors in patient lesion counts, which may have serious consequences in clinical trials. 
Moreover, DL methods typically produce predictors with deterministic outcomes. In contrast, traditional Bayesian machine learning provides not only a prediction, but also an uncertainty about it, through a probability density over outcomes. While mathematically principled, traditional Bayesian approaches to DL have not been widely used in applications due to implementation challenges and excessive training times. Recently, Gal and Ghahramani~\cite{gal2016dropout} presented a simpler approach to uncertainty estimation for DL, by training a dropout network and taking Monte Carlo (MC) samples of the prediction using dropout at test time. This approach produces an approximation of the posterior of the network's weights. In computer vision, modeling uncertainty improves the performance of a standard scene understanding network with no additional parameterization~\cite{kendall2015bayesian}. 
In the first application to medical image analysis, \cite{tanno2017bayesian}, add uncertainty modelling in a CNN to achieve state-of-the-art performance on the super-resolution of diffusion MR brain images, providing a discussion of how the predictive variance and MC sample variance uncertainties lead to this improvement.
In \cite{ozdemir2017propagating}, the authors leverage MC sample variance in a two-stage lung nodule detection system where they rely on uncertainty at nodule contours, and an initial prediction into the second stage, without an analysis of the usefulness of the uncertainty measure. In \cite{leibig2017leveraging}, the authors  perform an evaluation of the MC sample variance for image-based diabetic retinopathy diagnosis. They show that the sample variance is useful for referring cases to experts in this context. 

We present the first qualitative and quantitative comparison of the effectiveness of {\em several different uncertainty measures} derived from MC dropout in the context of DL for lesion segmentation and detection in medical images. 
We develop a 3D MS lesion segmentation CNN, augmented to provide four voxel-based uncertainty measures based on MC dropout: predictive variance, MC sample variance, predictive  entropy, and mutual information. The network is trained on a large proprietary, multi-scanner, multi-center, clinical trial dataset of patients with Relapsing-Remitting MS (RRMS). Voxel-wise uncertainties are combined to estimate lesion-level uncertainties. The resulting voxel-based segmentation and lesion-level detection performance are examined when choosing operating points on ROC curves (TPR vs. FDR) based on thresholded uncertainty levels.  Our results indicate that while bigger lesions have large voxel-based uncertainties primarily along the border, the smallest lesions have the highest lesion-level detection errors, along with the highest uncertainty. Selecting operating points based on dropout uncertainty measures proves to be a more robust and principled approach than typical thresholding based on the network's sigmoid output. 

\section{Proposed Method}
\begin{figure}
    \includegraphics[width=\textwidth]{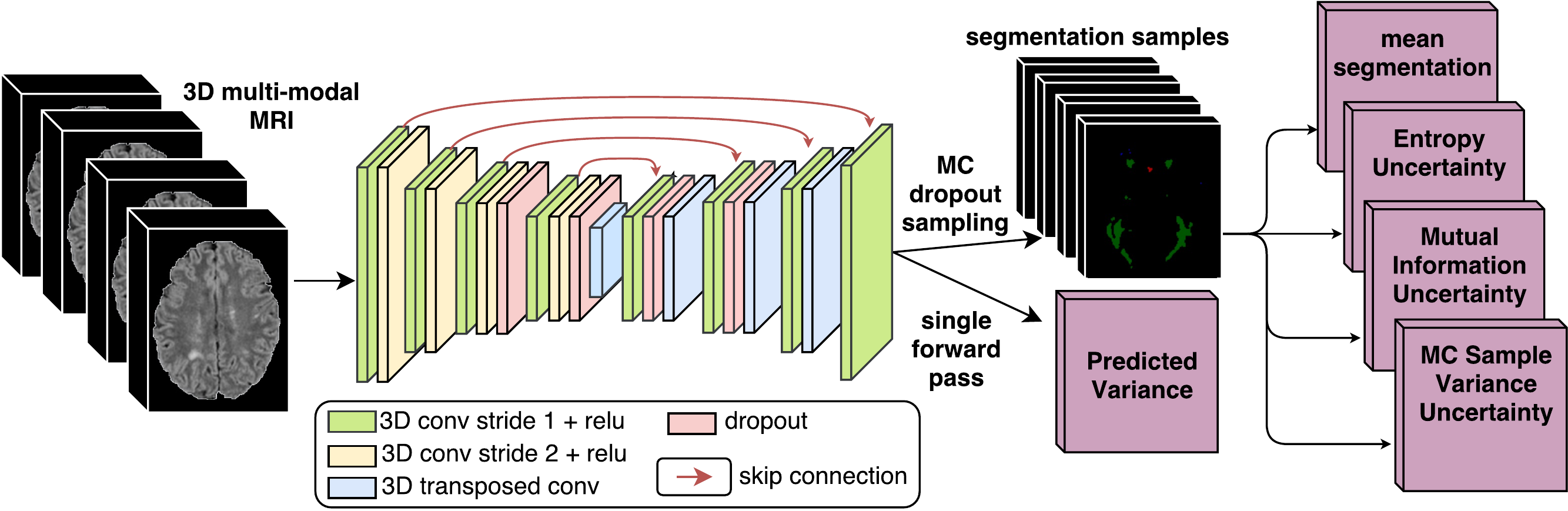}
    \caption{Network architecture. All convolutional operations are 3D and we use additive skip connections instead of concatenating ones to reduce the number of parameters}
 \label{fig:network}
\end{figure}

We develop a 3D fully convolutional neural network (CNN) with dropout to segment lesions from MRI sequences, by providing binary labels (lesion/non-lesion) to all voxels. The network is simultaneously trained to estimate uncertainty, as we will describe in detail below. Figure~\ref{fig:network} contains the flowchart of our method. During testing, we pass an unseen multi-modal MR volume, $x^*$, through the network $T$ times to obtain MC samples and estimate the uncertainty. The mean of the sample segmentations is used to obtain a single prediction. 

\subsection{Dropout as a Bayesian Approximation}
During training, pairs of multi-sequence 3D MRIs, ${\bf X}$, and their associated binary ground truth T2 lesion labels, ${\bf Y}$, are used to learn the weights, $W$, of the 3D CNN. 
To capture uncertainty in the model, a prior distribution is placed over $W$, and an estimate of the posterior $p(W|{\bf X}, {\bf Y})$ is computed.
Although computing this posterior analytically is intractable, variational methods can approximate it with a parameterized distribution $q(W)$ which minimizes the Kullback-Leibler (KL) divergence~\cite{blei2017variational}:
\begin{equation}\label{eq:kl_divergence}
 q^*(W) = \argmin_{q(W)} KL( q(W)\ ||\ p(W|{\bf X}, {\bf Y})).
\end{equation}
In \cite{gal2016dropout}, the authors show that minimizing the cross-entropy loss of a network with dropout applied after each layer of weights is equivalent to the minimization of the KL-divergence in Eq.~\eqref{eq:kl_divergence}, where the approximating distribution, $q(W)$, is a mixture of two Gaussians: \(\mathcal{N}(0, \epsilon\mathbf{I})\) and \(\mathcal{N}(M, \epsilon\mathbf{I})\) with small variances $\epsilon$, mixing coefficients \(p\) and \(1-p\) respectively, and where $M$ are the variational parameters for weights $W$. Their derivation treats sampling from the mixture of Gaussians as sampling from a Bernoulli random variable returning either $0$ or $M$, and links this to the application of dropout in a deep network.
This is a key result, and means that we can sample from a dropout network's outputs to approximate the  posterior $p(\hat{Y}|X^*,W)$ over the lesion label prediction $\hat{Y}$ for an input MRI $X^*$.

\subsection{Measures of Uncertainty in DL Networks}
We now describe the four uncertainty measures we will compute: prediction variance, which is learned directly from the training data and was discussed in~\cite{gal2016dropout}, and three stochastic sampling-based measure based on dropout: variance of MC samples, predictive entropy, and mutual information.

\subsubsection{Prediction Variance} During training, in addition the labels, the weights of the network are also trained to produce the prediction variance $\hat{V}$ at the output (Figure~\ref{fig:network}). DL networks for classification typically pass network outputs, $F_W$, through a sigmoid or softmax function to obtain predictions $\hat{Y}\in[0,1]$, which are then used in a loss function that compares them against ground truth labels $Y$. In our model, we follow the approach of~\cite{kendall2017uncertainties}, and assume the network outputs are corrupted by Gaussian noise with mean $0$, and variance $V$ at every voxel. The network is trained to output an estimate, $\hat{V}_W$, of the noise variance by reformulating prediction $\hat{Y}_W$ as:
\begin{equation}\label{eq:unary_plus_noise}
\hat{Y}_W = sigmoid(F_W + \mathcal{N}(0,{\bf I}\hat{V}_W)).
\end{equation}
During training, the Gaussian distribution is integrated out by taking $T$ MC samples of $\hat{Y}_W$ and $\hat{V}_W$. We then use the standard weighted, binary cross-entropy function, averaging across the MC samples. Because the prediction variance is used to compute segmentations $\hat{Y}$, and subsequently the cross-entropy loss, the weight updates to the network during backpropagation push the network to learn the variance estimates without having explicit labels for them.

\subsubsection{MC Sample Variance}
As in previous work applying MC dropout methods~\cite{kendall2015bayesian,leibig2017leveraging,tanno2017bayesian,ozdemir2017propagating}, the MC sample variance is a measure of uncertainty derived from the variance of the $T$ MC samples of the predicted segmentation, $var(\hat{Y}_1,...,\hat{Y}_T)$.

\subsubsection{Predictive Entropy} The predictive entropy is a measure of how much information is in the model predictive density function at each voxel $i$. We approximate the entropy for an input voxel $x_i^*$ across $T$ MC samples and $C$ classes with the following biased estimator~\cite{gal2017deep}:
\begin{equation}\label{eq:ent}
H[\hat{y}_i|x_i^*,{\bf X},{\bf Y}] \approx -\sum_{c=1}^{C} \frac{1}{T}\sum_{t=1}^{T} p(\hat{y}_i=c|x_i^*,W_t) log(\frac{1}{T}\sum_{t=1}^{T} p(\hat{y}_i=c|x_i^*,W_t)).
\end{equation}

\subsubsection{Mutual Information}
Finally, the mutual information between the model posterior density function and the prediction density function is approximated at each voxel $i$ as the difference between the entropy of the expected prediction, and the expectation of the model prediction entropies across samples~\cite{gal2017deep}:

\begin{equation}\label{eq:mutual_info}
MI[\hat{y}_i,W|x_i^*,{\bf X},{\bf Y}] \approx H[\hat{y}_i|x_i^*,{\bf X},{\bf Y}] - E[H[\hat{y}_i|x_i^*,W]]. 
\end{equation}

\subsection{Uncertainty-based Filtering in Lesion Segmentation/Detection}
The network outputs $\hat{y}_i$ computed as in Eq.\eqref{eq:unary_plus_noise} as well as the four defined measures of uncertainty $U_m(i),i=1\dots 4$ at every voxel $x_i^*$. The standard approach to generate a classification would be to compute the indicator function $\textbf{1}_{\hat{y}_i\geq \theta}$ where the threshold $\theta$ is specified (eg. $0.9$). When we use the uncertainty measure, we will additionally require that $U_m(i)$ is below another chosen threshold $\eta$ in order to produce the prediction.
If the predictions that are incorrect are uncertain, this filtering should increase the performance on remaining predictions.

In the context of neurological diseases such as MS, it important to perform lesion-level detection because changes in the a patient's lesion count are indicative of disease activity and progression. This requires a strategy to merge voxel-level uncertainty measures into lesion-level uncertainty, which is then used to perform lesion-wise filtering. 
Suppose we can generate a large set of candidate lesions. For a candidate $l$, composed of voxels $p,...,q$, we will compute the  lesion-uncertainty $U_m(l)$
from the voxel-wise uncertainties as: $U_m(l) = \sum_{i=p}^{q} log(U_m(i))$.
Taking the log-sum of the voxel-level uncertainties reflects the simplifying assumption that neighbouring voxels are conditionally independent, given that they are part of $l$. To make the uncertainties comparable through a single threshold value, we rescale the values $U_m(l)$ to $[0,1]$ by subtracting by the minimum lesion uncertainty and dividing by the range; we do this separately for each measure $m$. Detection is then performed using the uncertainty threshold and outputs in the same case as for the voxel-level. Further implementation details can be found in the supplementary material and code is available at \url{https://github.com/tanyanair/segmentation_uncertainty}.

\section{Experiments and Results}
The method was evaluated on a proprietary, multi-site, multi-scanner, clinical trial dataset of 1064 Relapsing-Remitting MS (RRMS) patients, scanned annually over a 24-month period. T1, T2, FLAIR, and PDW MRI sequences were acquired at a 1mm x 1mm x 3mm resolution and pre-processed with brain extraction~\cite{smith2002fast}, N3 bias field inhomogeneity correction~\cite{sled1998nonparametric}, Nyul image intensity normalization, and registration to the MNI-space. Ground truth T2 lesion segmentation masks were provided with the data. These were obtained using a proprietary approach where the result of an automated segmentation method was manually corrected by expert human annotators. The network was trained on 80\% of the subjects, with 10\% held out for validation and 10\% for testing (2182/251/251 scans for training/validation/testing respectively). We take 10 MC samples during training and testing for the evaluation of the uncertainties.

\begin{figure}
    \centering
        \includegraphics[width=\textwidth]{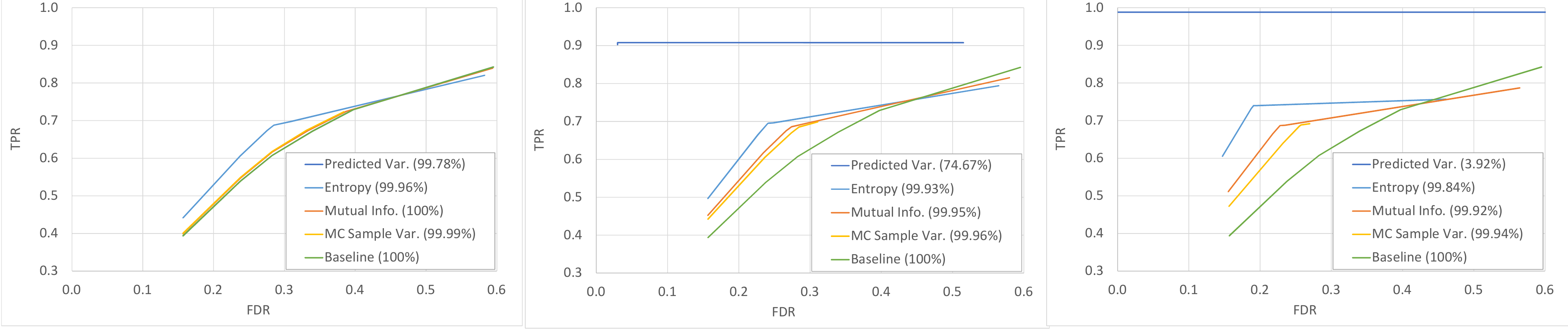}
        (a) voxel-wise thresholding at thresholds (left to right): 0.5, 0.1, 0.01\\ 
    \includegraphics[width=\textwidth]{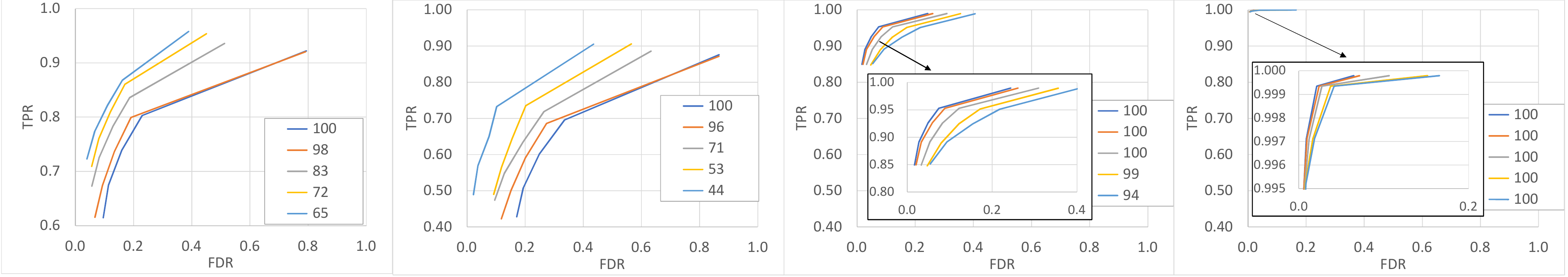}
        (b) lesion-wise thresholding across (left to right): all, small, medium, and large lesions\\
    \caption{FDR vs. TPR of retained predictions when (a) voxel-wise thresholding with each uncertainty measure and (b) lesion-wise thresholding with the entropy measure.
    In (a) the \% of voxels retained is provided with each uncertainty measure in the legend. In (b), the \% of lesions retained for each curve's uncertainty threshold is provided in the color coded legend for that plot. The uncertainty threshold used to generate a given color's curve in a plot is the same across these plots. Points along a curve correspond to different sigmoid thresholds used to binarize the model's segmentation output. Each plot contains a reference curve (100\%) corresponding to the model's baseline performance when no uncertainty thresholding is performed.}
    \label{fig:roc_all_lesion}
\end{figure}

To see if the uncertainty measures are useful and describe different information, we plot the voxel-level True Positive Rate ($TPR = \frac{TP}{TP+FN}$), and False Detection Rate ($FDR = 1-\frac{TP}{TP+FP}$) Receiver Operating Characteristic (ROC) curves for the retained voxels at different uncertainty thresholds, and them compare against a baseline ROC in which no uncertainty thresholding is performed (Fig.~\ref{fig:roc_all_lesion}a). At operating points of interest (FDR \textless\ 0.3), different measures lead to different percentage levels of voxel retention. The notably high TPR of the predicted variance curve can be attributed to the significantly lower voxel-level retention. 

To obtain lesion-level statistics from voxel segmentations, ground truth lesions smaller than 3 voxels were removed, as per clinical protocol. We performed voxel-level classification and candidate lesions are then obtained from lesion voxels by considering a surrounding, 18-connected neighbourhood in order to mitigate the impact of under-segmented ground truth in this dataset.
A true positive (TP) lesion is detected when the segmentation, including its 18-connected neighbourhood, overlaps with at least three, or more than 50\%, of the ground truth lesion voxels. Insufficient overlap results in a false negative (FN), and candidate lesions of 3 or more voxels that do not overlap with a ground truth lesion are counted as false positives (FP). Lesion-level TPR, and FDR ROC curves for retained predictions at different entropy thresholds are shown in Fig.~\ref{fig:roc_all_lesion}b. Results for the other measures can be found in the supplementary material, as they were extremely similar to one another.
At all operating points, across all measures, using uncertainty to exclude uncertain predictions improves performance on remaining
predictions, even when excluding just 2\% of the most uncertain lesions, due to the resulting reduction in the number of both FP and FN assertions.
In an analysis across small (3-10 vox), medium (11-50 vox), and large (51+) lesion bin sizes (Fig.~\ref{fig:roc_all_lesion}b), we find that using uncertainty to exclude predictions is helpful for small lesions regardless of the exact uncertainty measure used. This is because the model does not perform as well for small lesions, which constitute 40\% of lesions in the dataset, so removing the uncertain FP and FN segmentations improves the overall performance. However, for medium and large lesions, performance reduces slightly compared to non-thresholded segmentations because the DL model has very few FP's and FN's for medium and large lesions. Filtering out these sized lesions reduces TP's, reducing the performance for those sizes. 

\begin{figure}
    \centering
    \includegraphics[width=.8\textwidth]{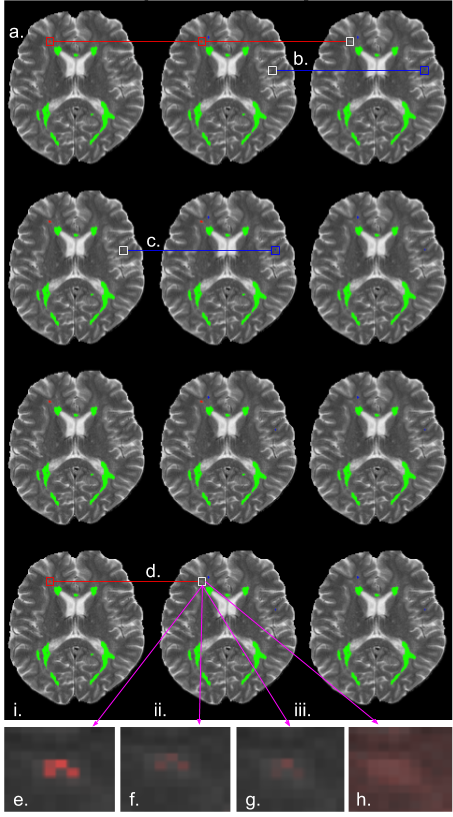}
    \caption{Visualization of a single MR slice's segmentation for different uncertainty thresholds. From top to bottom we use the following uncertainty measures: entropy, mutual information, MC sample variance, and predicted variance. From left to right we show \textbf{i.}~Baseline (no uncertainty thresholding) lesion segmentation. \textbf{ii., iii.}~Segmentation at two different thresholds for the uncertainty corresponding to that row (ie. more lesions are excluded in iii.). Lesions are coloured with the following scheme: TP: green, FP:red, FN: blue, and True Negative (TN): white. Lesions~\textbf{a, d} show the uncertainty corresponding to a FP in the baseline segmentation turned  into a TN as the uncertainty threshold is increased. \textbf{b, c}~show a TP (from another slice) becoming a FN. \textbf{d}~shows a FP (from another slice) becoming a TN. The lesion uncertainties at~\textbf{d} are shown for~\textbf{e} entropy, \textbf{f} mutual information, \textbf{g} MC sample variance, and \textbf{h} predicted variance. Note that the threshold to turn lesions into TN's or FN's is different for different uncertainty measures. All MRI are courtesy of Neurorx Research.}
    \label{fig:prediction_across_unc2}
\end{figure}

Fig.~\ref{fig:prediction_across_unc2}(e-h) provides an example of uncertainties themselves. In general, measures computed from stochastic dropout samples are more uncertain around lesion contours. Relative to other work~\cite{ozdemir2017propagating,tanno2017bayesian}, the MC sample variance is very small, even around lesion contours, but mutual information and predictive entropy reflect the boundary-uncertainty more intensely. The learned, predictive variance reflects data uncertainty throughout contours in the entire MRI (e.g. boundaries between white matter and grey matter). 
Despite these voxel-wise differences, when accumulating evidence to the lesion-level, the different measures tend to rank lesions is the same order of certainty, albeit on different scales, which leads to filtering out the same lesions, at different thresholds (ie. Fig.~\ref{fig:prediction_across_unc2}c-d)). 
One interpretation is that between taking MC dropout samples, and computing the uncertainty measures, no new information is added.
We also note that small lesions are relatively more uncertain than medium and large lesions. This is a consequence of computing lesion uncertainty from the log sum of all the uncertainty values in a detected lesion area. Although large lesions have larger, more uncertain contours, the accumulation of lesion-evidence within the boundary provides an overwhelming certainty that there is a lesion there. This is not the case for small lesions (less evidence).

\section{Conclusion}
We developed a 3D MS lesion segmentation CNN, augmented to provide four voxel-based uncertainties, and showed how these can be accumulated to estimate lesion-level uncertainties. 
Our results indicate that filtering based on uncertainty
greatly improves lesion detection accuracy for small lesions, which make up 40\% of the dataset, indicating that high uncertainty does indeed reflect incorrect predictions.
Moreover, uncertainty measures in the results of an automatic, DL detection or segmentation method provide clinicians or radiologists with information permitting them to quickly assess whether to accept or reject lesions of high uncertainty, for example, or further analyze uncertain lesion boundaries. This could facilitate the wider adoption of DL methods into clinical work-flows. 

\subsubsection*{Acknowledgements.}
This work was supported by the Canadian NSERC Discovery and CREATE grants, and an award from the International Progressive MS Alliance (PA-1603-08175).


\bibliographystyle{splncs03}
\bibliography{\jobname}

\begin{thebibliography}{10}
\providecommand{\url}[1]{\texttt{#1}}
\providecommand{\urlprefix}{URL }

\bibitem{blei2017variational}
Blei, D.M., et~al.: Variational inference. ASA  112(518),  859--877 (2017)

\bibitem{brosch2016deep}
Brosch, T., et~al.: Deep 3{D} convolutional encoder networks with shortcuts for
  multiscale feature integration applied to {M}ultiple {S}clerosis lesion
  segmentation. TMI  35(5),  1229--1239 (2016)

\bibitem{carass2017longitudinal}
Carass, A., et~al.: Longitudinal {M}ultiple {S}clerosis lesion segmentation:
  Resource and challenge. NeuroIm.  148,  77--102 (2017)

\bibitem{gal2016dropout}
Gal, Y., Ghahramani, Z.: Dropout as a bayesian approximation: Representing
  model uncertainty in deep learning. ICML pp. 1050--1059 (2016)

\bibitem{gal2017deep}
Gal, Y., et~al.: Deep bayesian active learning with image data. ICML  (2017)

\bibitem{ghafoorian2016non}
Ghafoorian, M., et~al.: Non-uniform patch sampling with deep convolutional
  neural networks for white matter hyperintensity segmentation. ISBI pp.
  1414--1417 (2016)

\bibitem{hernandez2016team}
Hernandez, C., et~al.: Team {D}elft\text{'s} robot winner of the {A}mazon
  {P}icking {C}hallenge 2016. Robot World Cup  (2016)

\bibitem{kendall2017uncertainties}
Kendall, A., Gal, Y.: What uncertainties do we need in bayesian deep learning
  for computer vision? NIPS pp. 5580--5590 (2017)

\bibitem{kendall2015bayesian}
Kendall, A., et~al.: Bayesian {S}eg{N}et: {M}odel uncertainty in deep
  convolutional encoder-decoder architectures for scene understanding. BMVC
  (2017)

\bibitem{leibig2017leveraging}
Leibig, C., et~al.: Leveraging uncertainty information from deep neural
  networks for disease detection. Nature  7(1),  17816 (2017)

\bibitem{menze2015multimodal}
Menze, B.H., et~al.: The multimodal brain tumor image segmentation benchmark
  (brats). TMI  34(10),  1993--2024 (2015)

\bibitem{ozdemir2017propagating}
Ozdemir, O., et~al.: Propagating uncertainty in multi-stage bayesian
  convolutional neural networks with application to pulmonary nodule detection.
  NIPS  (2017)

\bibitem{russakovsky2015imagenet}
Russakovsky, O., et~al.: Imagenet large scale visual recognition challenge.
  IJCV  115(3),  211--252 (2015)

\bibitem{sled1998nonparametric}
Sled, J.G., et~al.: A nonparametric method for automatic correction of
  intensity nonuniformity in {MRI} data. TMI  17(1),  87--97 (1998)

\bibitem{smith2002fast}
Smith, S.M.: Fast robust automated brain extraction. HBM  17(3),  143--155
  (2002)

\bibitem{styner20083d}
Styner, M., et~al.: 3{D} segmentation in the clinic: A grand challenge {II}:
  M{S} lesion segmentation. MIDAS  2008,  1--6 (2008)

\bibitem{tanno2017bayesian}
Tanno, R., et~al.: Bayesian image quality transfer with cnns: Exploring
  uncertainty in dmri super-resolution. In: MICCAI. pp. 611--619. Springer
  (2017)

\end{thebibliography}
\end{document}